\documentclass[11pt]{article}
\usepackage[margin=1in]{geometry}
\usepackage{amsmath,amssymb}
\usepackage{graphicx}
\usepackage{authblk}
\usepackage{hyperref}
\usepackage{cite}
\usepackage{algorithm}
\usepackage{algorithmic}
\usepackage{amsthm}  
\newtheorem{theorem}{Theorem}[section]
\newtheorem{proposition}[theorem]{Proposition}
\usepackage{placeins}  

\title{AMSFL: Adaptive Multi-Step Federated Learning via Gradient Difference-Based Error Modeling}

\author[1]{Ganglou Xu}
\affil[1]{College of Computer Science and Technology, Hainan University, Haikou, China}
\affil[ ]{\texttt{13056960805@163.com}}

\date{}

\begin{document}

\maketitle

\begin{abstract}
Federated learning faces critical challenges in balancing communication efficiency and model accuracy. One key issue lies in the approximation of update errors without incurring high computational costs. In this paper, we propose a lightweight yet effective method called Gradient Difference Approximation (GDA), which leverages first-order information to estimate local error trends without computing the full Hessian matrix. The proposed method forms a key component of the Adaptive Multi-Step Federated Learning (AMSFL) framework and provides a unified error modeling strategy for large-scale multi-step adaptive training environments.
\end{abstract}

\section{Introduction}

Federated learning (FL) has emerged as a promising paradigm that enables distributed model training across multiple clients without centralising raw data~\cite{mcmahan2017communication}. It preserves data privacy while leveraging data diversity across devices. However, in real-world deployments involving edge devices or heterogeneous clients, federated learning faces several fundamental challenges such as non-i.i.d. data distribution, resource heterogeneity, and unstable convergence due to delayed or inconsistent updates.

Among these, one critical but often under-addressed issue is the cumulative error propagation introduced by multiple local updates. In typical FL settings such as FedAvg, clients perform several local gradient descent steps before communicating with the server. While this strategy reduces communication overhead, it also introduces deviation between the local and global models, leading to potential convergence degradation, especially in highly heterogeneous systems.

Traditional solutions often involve second-order information, such as using the Hessian matrix to model curvature and guide aggregation. Although theoretically sound, such approaches are computationally intensive and impractical on low-resource devices. To overcome this, there is a growing demand for lightweight yet principled error modeling techniques that enable effective control of local update deviations with minimal overhead.

\textbf{This paper proposes a first-order approximation technique, termed Gradient Difference Approximation (GDA)}, to estimate local update errors without computing Hessians. This method forms the foundation of a broader framework called Adaptive Multi-Step Federated Learning (AMSFL), where clients dynamically determine their optimal number of local steps by jointly considering update efficiency and error propagation. The key insight is to express error trends using gradient differences between successive models, enabling efficient and scalable approximation of second-order effects.

Furthermore, we formulate the multi-step optimization problem as a constrained error minimization under device-specific resource budgets, and derive theoretical guarantees on convergence and error bounds under mild assumptions. Extensive experiments on standard benchmarks demonstrate that our method achieves better trade-offs between communication efficiency and model performance compared to existing approaches.
\section{Related Work}

\subsection{Error Modeling in Federated Learning}

Error propagation is a central concern in federated optimization. In classical FedAvg~\cite{mcmahan2017communication}, clients perform multiple local updates before communicating with the server, leading to discrepancies between the local and global models. This discrepancy can cause delayed convergence or divergence, especially when client data are non-i.i.d. or heterogeneous in scale.

To address this, FedProx~\cite{li2020federated} introduced a proximal term to penalize local deviation, effectively regularizing local updates. SCAFFOLD~\cite{karimireddy2020scaffold} introduced control variates to reduce the client drift in heterogeneous settings. Other works such as FedDyn~\cite{acar2021federated} and Mime~\cite{karimireddy2021mime} incorporated curvature-related feedback or dynamic regularization to counterbalance local updates.

However, most of these approaches either assume bounded dissimilarity across clients or rely on global regularization, without providing fine-grained estimates of how local update steps quantitatively affect global error accumulation. Furthermore, they often overlook the impact of step size and update count on convergence, especially under practical constraints such as limited computation budgets or varying local resources.

\subsection{First-Order Approximation Methods}

Second-order methods (e.g., involving Hessian matrices) provide a more precise view of curvature and error growth, but are computationally intractable for large models or edge devices. As an alternative, first-order approximation methods have been extensively explored in distributed optimization~\cite{zhang2015disco,shamir2014communication}. These approaches avoid the complexity of second-order information by relying on gradient differences, residuals, or momentum-based corrections.

In federated settings, some methods implicitly leverage first-order information to adjust updates. For instance, FedDyn approximates Hessian effects via a dynamic regularization term. Yet, few works explicitly construct gradient-based approximations of Hessian-vector products as a principled means to analyze or control error propagation.

Our proposed Gradient Difference Approximation (GDA) fills this gap by directly using first-order gradient variations to replace second-order terms in Taylor expansions. This enables error tracking and convergence control with linear time complexity, making it suitable for resource-constrained federated environments.

\subsection{Multi-Step Local Training and Step Control Strategies}

Local training with multiple steps between communications reduces transmission cost but introduces potential instability. Several studies have investigated how the number of local updates affects convergence. Works such as Local SGD~\cite{stich2018local} and FedNova~\cite{wang2020tackling} analyze the trade-off between local computation and global synchronization frequency.

While many algorithms fix the number of local steps (e.g., 5 or 10), recent approaches aim to make this number adaptive. FedAdapt~\cite{wang2020optimizing} proposes an online tuning framework to adjust local steps per round. However, it lacks a formal connection between local steps and global error behavior. Other heuristics attempt to adjust local updates based on loss curvature or gradient norm, but few provide theoretical guarantees.

Our work builds on this line by providing a rigorous error formulation that explicitly links the number of local steps to global error bounds. By solving a constrained optimization problem, we derive client-specific adaptive step sizes that improve both convergence and efficiency.

\vspace{0.5em}
\noindent
\textbf{Our contributions are summarised as follows:}
\begin{itemize}
    \item We propose \textbf{Gradient Difference Approximation (GDA)}, a first-order method to approximate second-order model deviation effects using only gradient differences, with negligible computational cost.
    
    \item We derive a novel \textbf{error propagation formulation} for multi-step local updates, and provide theoretical \textbf{upper bounds on global model deviation} under realistic assumptions.
    
    \item We formulate and solve the \textbf{multi-step optimization problem} under time and cost constraints, leading to an adaptive training framework called AMSFL with provable convergence guarantees.
    
    \item We conduct comprehensive experiments showing that our method improves convergence efficiency and accuracy while significantly reducing communication and computation overhead.
\end{itemize}
\section{Methodology}

\subsection{Problem Formulation and Notation}

We consider a federated learning system consisting of $N$ clients collaboratively training a shared global model $\mathbf{w} \in \mathbb{R}^d$ under a central server. Each client $i \in \{1, 2, \dots, N\}$ holds a local dataset $\mathcal{D}_i$ and optimizes a local loss function $F_i(\mathbf{w})$, which is defined as the empirical risk over $\mathcal{D}_i$:
\begin{equation}
    F_i(\mathbf{w}) = \frac{1}{|\mathcal{D}_i|} \sum_{(x, y) \in \mathcal{D}_i} \ell(\mathbf{w}; x, y),
\end{equation}
where $\ell(\mathbf{w}; x, y)$ is a sample-wise loss function (e.g., cross-entropy).

The global objective is to minimize the weighted average loss:
\begin{equation}
    F(\mathbf{w}) = \sum_{i=1}^N p_i F_i(\mathbf{w}), \quad \text{where } p_i = \frac{|\mathcal{D}_i|}{\sum_{j=1}^N |\mathcal{D}_j|}.
\end{equation}

Federated optimization proceeds in communication rounds indexed by $k$. In each round:
\begin{enumerate}
    \item The server broadcasts the current global model $\mathbf{w}^{(k)}$ to a selected subset of clients.
    \item Each participating client performs $m_i^{(k)}$ local updates using stochastic gradient descent (SGD) on $F_i$, starting from $\mathbf{w}^{(k)}$.
    \item The updated local models $\mathbf{w}_i^{(k)}$ are sent back and aggregated by the server to form the new global model.
\end{enumerate}

Let $\mathbf{w}_i^{(k)}$ denote the model held by client $i$ after $m_i^{(k)}$ local SGD steps in round $k$, with learning rate $\eta > 0$. The local update rule is:
\begin{equation}
    \mathbf{w}_{i}^{(k)} = \mathbf{w}^{(k)} - \eta \sum_{j=1}^{m_i^{(k)}} \nabla F_i(\mathbf{w}_{i, j}^{(k)}),
\end{equation}
where $\mathbf{w}_{i, j}^{(k)}$ is the intermediate model at step $j$ of client $i$'s local update.

We define the \textbf{local model deviation} of client $i$ in round $k$ as:
\begin{equation}
    \delta_i^{(k)} := \mathbf{w}_i^{(k)} - \mathbf{w}^{(k)}.
\end{equation}

This deviation introduces an error in the aggregation step. To quantify this, the global model update becomes:
\begin{equation}
    \mathbf{w}^{(k+1)} = \sum_{i=1}^N p_i \mathbf{w}_i^{(k)} = \mathbf{w}^{(k)} + \sum_{i=1}^N p_i \delta_i^{(k)}.
\end{equation}

Thus, the effectiveness of global optimization is closely tied to the magnitude and direction of the deviations $\delta_i^{(k)}$, which are influenced by the local steps $m_i^{(k)}$, data heterogeneity, and gradient dynamics. Controlling this error accumulation is the primary challenge addressed in this work.

To this end, we propose a first-order approximation mechanism—Gradient Difference Approximation (GDA)—to model $\delta_i^{(k)}$-induced errors and design a multi-step control strategy to minimize their negative effects while respecting client resource budgets.
\subsection{Error Modeling via Gradient Difference Approximation (GDA)}
\label{sec:error_modeling}

To accurately model how local updates affect global optimization, we analyze the propagation of model error across communication rounds. Let $w^*$ be the optimal model that minimizes the global objective $F(w)$, and define the global error at round $k$ as:
\[
e^{(k)} := \mathbf{w}^{(k)} - \mathbf{w}^*.
\]

During communication round $k$, client $i$ initializes with the global model $\mathbf{w}^{(k)}$ and performs $t_i$ local SGD updates with step size $\eta$. Denote by $\mathbf{w}_i^{(t_i)}$ the resulting local model after $t_i$ steps, and define the local error:
\[
e_i^{(t_i)} := \mathbf{w}_i^{(t_i)} - \mathbf{w}^*.
\]

We now track the evolution of this local error. Let $\Delta g_i^{(t)} := \nabla F_i(\mathbf{w}_{i,t}) - \nabla F_i(\mathbf{w}^{(k)})$ be the gradient deviation at local step $t$. Then we define the accumulated gradient deviation:
\begin{equation}
\Delta_i^{(t_i)} := \sum_{t=0}^{t_i - 1} \sum_{j=1}^{t} \Delta g_i^{(j)}.
\label{eq:delta_def}
\end{equation}

Then, assuming constant step size and fixed starting point $\mathbf{w}^{(k)}$, the local error after $t_i$ updates becomes:
\begin{equation}
e_i^{(t_i)} = e_i^{(0)} - \eta t_i \nabla F_i(\mathbf{w}^{(k)}) - \eta \Delta_i^{(t_i)},
\label{eq:local_error}
\end{equation}
where $e_i^{(0)} = \mathbf{w}^{(k)} - \mathbf{w}^*$.

After aggregation, the new global model is:
\[
\mathbf{w}^{(k+1)} = \sum_i \omega_i \mathbf{w}_i^{(t_i)},
\]
where $\omega_i$ are aggregation weights (e.g., proportional to data size). The global error becomes:
\begin{align}
e^{(k+1)} &= \mathbf{w}^{(k+1)} - \mathbf{w}^* 
= \sum_i \omega_i e_i^{(t_i)} \nonumber \\
&= e^{(k)} - \eta \sum_i \omega_i t_i \nabla F_i(\mathbf{w}^{(k)}) - \eta \sum_i \omega_i \Delta_i^{(t_i)}.
\label{eq:global_error}
\end{align}

\paragraph{Interpretation.}  
This formulation shows that the global error at round $k+1$ results from:
\begin{itemize}
    \item the residual error from the previous round $e^{(k)}$;
    \item the descent direction formed by the weighted sum of local gradients;
    \item an accumulated error term $\Delta_i^{(t_i)}$ representing curvature effects and gradient drift during local training.
\end{itemize}

We will use this error recursion as the foundation for analyzing convergence behaviour in the next section, as well as for deriving upper bounds and step-size optimization in AMSFL.
\subsection{Theoretical Results}
\label{sec:theory}

We now present the theoretical analysis of our AMSFL algorithm under the error recursion derived in Eq.~\eqref{eq:global_error}. We provide a convergence bound under smoothness assumptions and characterize the impact of multi-step local updates on error dynamics and residual stability.

\vspace{0.5em}
\paragraph{Assumptions.}
We make the following standard assumptions:

\begin{itemize}
    \item[] \textbf{(A1) $L$-smoothness:} Each $F_i(\cdot)$ is $L$-smooth: $\|\nabla F_i(w) - \nabla F_i(w')\| \leq L\|w - w'\|$.
    \item[] \textbf{(A2) Bounded gradient norm:} $\|\nabla F_i(w)\| \leq G$ for all $w$.
    \item[] \textbf{(A3) Gradient difference approximation (GDA):}
    \[
    \left\|\nabla^2 F_i(w) \cdot \delta - \left( \nabla F_i(w + \delta) - \nabla F_i(w) \right)\right\| \leq \frac{L}{2} \|\delta\|^2.
    \]
    \item[] \textbf{(A4) Local drift accumulation:} For client $i$ with $t_i$ local steps,
    \[
    \left\| \Delta_i^{(t_i)} \right\| \leq \frac{L G}{2} t_i (t_i - 1).
    \]
\end{itemize}

We denote the following aggregated quantities:
\[
E := \sum_i \omega_i t_i, \quad
D_k^2 := \sum_i \omega_i \cdot \frac{t_i(t_i - 1)}{2}, \quad
\Delta_k := \eta^2 G^2 E^2 + \eta^2 L^2 G^2 D_k^2.
\]

\begin{theorem}[Error Recursion of AMSFL]
\label{thm:recursion}
Under assumptions (A1)–(A4), the global error across communication rounds satisfies:
\begin{equation}
\label{eq:recursion}
\|e^{(k+1)}\|^2 \leq \|e^{(k)}\|^2 - 2\eta E \left\langle \nabla F(w^{(k)}), e^{(k)} \right\rangle + \Delta_k.
\end{equation}
\textit{Interpretation:} The update consists of a descent term driven by the weighted global gradient, and a residual error determined by local training steps and curvature-induced drift.
\end{theorem}

\begin{theorem}[Linearized Form and Residual Convergence]
\label{thm:residual_bound}
Using the inequality $2ab \leq \theta a^2 + \frac{1}{\theta} b^2$, we linearize Eq.~\eqref{eq:recursion}:
\[
\|e^{(k+1)}\|^2 \leq (1 - \theta)\|e^{(k)}\|^2 + \left(1 + \frac{1}{\theta}\right) \Delta_k,
\]
for any $\theta \in (0, 1)$. This leads to the residual error bound:
\[
\limsup_{k \to \infty} \|w^{(k)} - w^*\|^2 \leq \left(1 + \frac{1}{\theta}\right) \Delta_k.
\]
\textit{Interpretation:} AMSFL converges to a bounded error region depending on $\eta$, $t_i$, and data heterogeneity. Smaller step sizes and fewer local updates reduce residual error.
\end{theorem}

\begin{proposition}[Approximation Error of GDA]
\label{prop:gda_error}
Let $F_i$ be twice differentiable and $L$-smooth. Then for any update $\delta$:
\[
\left\| \nabla^2 F_i(w) \cdot \delta - (\nabla F_i(w+\delta) - \nabla F_i(w)) \right\| \leq \frac{L}{2} \|\delta\|^2.
\]
\textit{Interpretation:} The gradient difference approximation introduces second-order bounded error and justifies the use of GDA in resource-constrained federated optimization.
\end{proposition}

\begin{theorem}[Optimal Local Step Allocation under Budget Constraint]
\label{thm:adaptive}
Assume each client $i$ incurs a cost $C_i$ per step and the total time budget is $T$. Then the following optimization problem:
\[
\min_{\{t_i\}} \Delta_k \quad \text{s.t.} \quad \sum_i \omega_i C_i t_i \leq T
\]
has optimal solution:
\[
t_i^* \propto \left(\frac{1}{C_i}\right)^{1/2}, \quad \text{when } \Delta_i^{(t_i)} \sim t_i^2.
\]
\textit{Interpretation:} More steps are assigned to lower-cost clients, balancing speed and drift. The sublinear allocation mitigates error amplification from excessive local updates.
\end{theorem}

\subsection{Adaptive Step Scheduling under Time Constraint}
\label{sec:step_optimization}

In real federated environments, training must respect hardware and latency constraints. Hence, local step allocation must obey a strict wall-clock time budget per communication round. To optimize training efficiency while minimizing the global model error, we design a time-aware adaptive scheduling strategy based on the error structure in Theorem~\ref{thm:residual_bound}.

\paragraph{Objective.}
From Theorem~\ref{thm:residual_bound}, the dominant residual term at round $k$ is:
\[
\Delta_k = \eta^2 G^2 E^2 + \eta^2 L^2 G^2 D_k^2,
\]
where \( E = \sum_i \omega_i t_i \), \( D_k^2 = \sum_i \omega_i \cdot \frac{t_i(t_i - 1)}{2} \). Therefore, we define the total error cost as:
\begin{equation}
\min_{\{t_i\}} \quad \alpha \sum_i \omega_i t_i + \beta \sum_i \omega_i \cdot \frac{t_i(t_i - 1)}{2},
\label{eq:step_obj}
\end{equation}
where \( \alpha = 2\eta\sqrt{\mu} G_k \), \( \beta = \frac{1}{2} \eta^2 L^2 G^2 \).

\paragraph{Time Constraint.}
Let each client $i$ require $c_i$ seconds per local step and incur $b_i$ seconds of communication delay. Let $S$ denote the total time budget for the round. Then the scheduling must satisfy the constraint:
\[
\sum_i (c_i t_i + b_i) \leq S, \quad \text{where } t_i \in \mathbb{N}^+, \forall i.
\]

\paragraph{Optimization Problem.}
We obtain the following integer program:
\begin{equation}
\begin{aligned}
\min_{\{t_i\}} & \quad \alpha \sum_i \omega_i t_i + \beta \sum_i \omega_i \cdot \frac{t_i(t_i - 1)}{2} \\
\text{s.t.} & \quad \sum_i (c_i t_i + b_i) \leq S, \quad t_i \in \mathbb{N}^+.
\end{aligned}
\label{eq:step_opt}
\end{equation}

\paragraph{Greedy Scheduling Algorithm.}
To solve the nonlinear constrained problem efficiently, we propose a greedy heuristic. Starting with $t_i = 1$ for all clients (minimum participation), we iteratively assign one additional step to the client with the least cost-to-error ratio until the time budget is exhausted.

\begin{algorithm}[H]
\caption{Greedy Adaptive Step Assignment under Time Budget}
\label{alg:greedy_step}
\begin{algorithmic}[1]
\STATE \textbf{Input:} weights $\{\omega_i\}$, step costs $\{c_i\}$, communication delays $\{b_i\}$, time budget $S$, constants $\alpha$, $\beta$
\STATE Initialize $t_i \gets 1$ for all $i$, total time $T \gets \sum_i (c_i + b_i)$
\WHILE{$T < S$}
    \FOR{each client $i$}
        \STATE Compute incremental cost: $\Delta_i = \frac{\alpha \omega_i + \beta \omega_i (2 t_i - 1)/2}{c_i}$
    \ENDFOR
    \STATE Let $j = \arg\min_i \Delta_i$
    \STATE $t_j \gets t_j + 1$, \quad $T \gets T + c_j$
\ENDWHILE
\STATE \textbf{Return:} $\{t_i\}$
\end{algorithmic}
\end{algorithm}

\paragraph{Discussion.}
This adaptive strategy ensures that clients with low computation cost and small error amplification potential are assigned more steps. Compared to fixed-step baselines (e.g., FedAvg), our method dynamically adapts to time constraints while minimizing residual error, which is crucial for practical deployment in resource-heterogeneous environments.
\section{Theoretical Analysis}
\label{sec:theory_main}

This section presents the theoretical guarantees of the proposed AMSFL algorithm. We focus on quantifying the effects of gradient difference approximation and multi-step local training under time-constrained resource settings.

We formally state four key results:

\begin{itemize}
    \item \textbf{Theorem 3.1:} Error propagation recurrence of the global model;
    \item \textbf{Theorem 3.2–3.3:} Convergence and residual error upper bounds of AMSFL;
    \item \textbf{Proposition 3.3:} Approximation error of the gradient difference (GDA);
    \item \textbf{Theorem 3.4:} Optimal adaptive step allocation under time budget constraint.
\end{itemize}

\noindent
Due to the technical length of these results, all formal proofs are deferred to \textbf{Appendix~\ref{appendix:proofs}}.
\section{Experimental Setup and Results}
\label{sec:experiment}

\subsection{Experimental Setup}

\subsubsection{Platform and Dataset}
We conduct experiments on the NSL-KDD dataset, a widely used benchmark for network intrusion detection. The data is partitioned into five clients under non-IID conditions to simulate real-world heterogeneity. All clients train a consistent model using SGD. Experiments are executed on a workstation with an Intel Xeon CPU and 128GB RAM.

\subsubsection{Compared Methods}
We compare our proposed \textbf{AMSFL} against six state-of-the-art federated learning methods: FedAvg~\cite{mcmahan2017communication}, FedProx~\cite{li2020federated}, FedNova~\cite{wang2020tackling}, SCAFFOLD~\cite{karimireddy2020scaffold}, FedDyn~\cite{acar2021federated}, and FedCSDA~\cite{altomare2024fedcsda}.

\subsubsection{Evaluation Metrics}
We evaluate the following aspects: (1) client-wise and global accuracy, (2) average training time per round, (3) convergence time and rounds to 89\% accuracy, and (4) stability across multiple runs.

\subsection{Experimental Results}

\subsubsection{Accuracy and Efficiency Comparison}

Table~\ref{tab:main_result} presents the accuracy across five clients and global performance, as well as the average training time per round. AMSFL achieves the highest global accuracy (0.9023) and exhibits strong client-level consistency. Moreover, AMSFL completes each training round in only 0.58 seconds, significantly faster than FedProx (1.02s) and SCAFFOLD (1.11s).

\begin{table}[htbp]
\centering
\caption{Accuracy and Average Runtime Comparison under 100s Training Budget}
\label{tab:main_result}
\begin{tabular}{l|ccccc|cc}
\hline
\textbf{Method} & $ACC_{C_1}$ & $ACC_{C_2}$ & $ACC_{C_3}$ & $ACC_{C_4}$ & $ACC_{C_5}$ & $ACC_{\text{global}}$ & Time/Round \\
\hline
FedAvg   & 0.8796 & 0.8849 & 0.8554 & 0.8535 & 0.8754 & 0.8987 & 0.85 \\
SCAFFOLD & 0.8783 & 0.8829 & 0.8527 & 0.8501 & 0.8725 & 0.8961 & 1.11 \\
FedProx  & 0.8805 & \textbf{0.8846} & 0.8548 & 0.8521 & 0.8730 & 0.8954 & 1.01 \\
FedNova  & 0.8753 & 0.8749 & 0.8481 & 0.8469 & 0.8632 & 0.8850 & 1.05 \\
FedDyn   & 0.8825 & 0.8859 & 0.8565 & 0.8572 & 0.8765 & 0.9010 & 0.83 \\
FedCSDA  & 0.8742 & 0.8805 & 0.8485 & 0.8460 & 0.8654 & 0.8763 & 1.02 \\
\textbf{AMSFL} & \textbf{0.8864} & 0.8834 & \textbf{0.8907} & \textbf{0.8659} & \textbf{0.8949} & \textbf{0.9023} & \textbf{0.58} \\
\hline
\FloatBarrier
\end{tabular}
\end{table}

\subsubsection{Convergence Speed}

Table~\ref{tab:convergence} shows the number of communication rounds and total time required to reach the target accuracy of 89\%. AMSFL converges in just 49.03 seconds, outperforming FedAvg (54.56s), FedDyn (68.44s), and FedCSDA (138s), despite using more communication rounds.

\begin{table}[htbp]
\centering
\caption{Convergence Time and Rounds to Reach 89\% Accuracy}
\label{tab:convergence}
\begin{tabular}{l|c|c|c|c}
\hline
\textbf{Method} & Target Accuracy & Comm. Time (s) & Comm. Rounds & Time/Round (s) \\
\hline
FedAvg   & 0.89 & 54.56 & 13 & 4.20 \\
SCAFFOLD & 0.89 & 84.15 & 16 & 5.26 \\
FedProx  & 0.89 & 80.45 & 16 & 5.03 \\
FedNova  & 0.89 & 117.05 & 23 & 5.08 \\
FedDyn   & 0.89 & 68.44 & \textbf{15} & 4.56 \\
FedCSDA  & 0.89 & 138.00 & 26 & 5.31 \\
\textbf{AMSFL} & \textbf{0.89} & \textbf{49.03} & \textbf{23} & \textbf{2.13} \\
\hline

\end{tabular}
\end{table}

\subsubsection{Stability and Generalization}

\begin{figure}[htbp]
\centering
\includegraphics[width=0.8\linewidth]{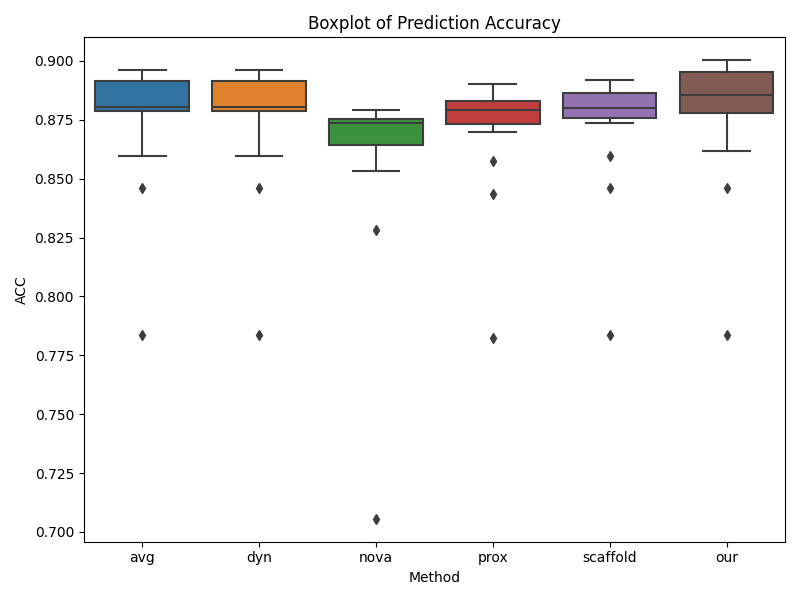} \\
\small{Figure A: Boxplot of Prediction Accuracy} \\
\vspace{1em}
\includegraphics[width=0.8\linewidth]{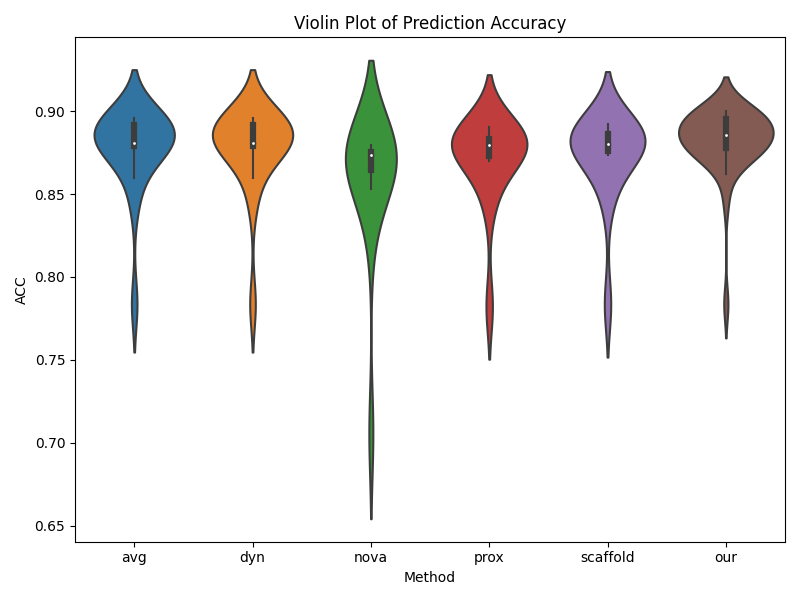} \\
\small{Figure B: Violin Plot of Prediction Accuracy}
\caption{Accuracy Distribution across 50 Trials for Each Method}
\label{fig:acc_violin}
\end{figure}

Figures~\ref{fig:acc_violin}(A) and~(B) visualize the accuracy distributions over 50 independent training runs. AMSFL demonstrates both high median accuracy and minimal variance, confirming its robustness against randomness and client variability.

\FloatBarrier

\subsection{Discussion}

The experimental results confirm that AMSFL provides a superior balance of accuracy, efficiency, and stability. Its gradient difference approximation effectively models local update drift, while the adaptive step allocation optimizes training within limited budgets. Compared with state-of-the-art baselines, AMSFL not only achieves the highest accuracy but also converges faster and generalizes better across diverse client distributions. These findings strongly support the applicability of AMSFL in real-world federated learning deployments.

\section{Conclusion and Future Work}
\label{sec:conclusion}
In this work, we propose AMSFL, an adaptive multi-step federated learning framework that incorporates gradient difference approximation to explicitly model local training drift and communication heterogeneity. We establish a theoretical foundation for the error propagation dynamics under GDA-based local updates and derive convergence guarantees under fixed and adaptive step-size regimes. Empirical results on the NSL-KDD dataset demonstrate that AMSFL achieves superior accuracy, faster convergence, and improved stability compared to existing baselines, validating both the theoretical insights and the effectiveness of adaptive scheduling.

Looking forward, several directions remain open for future investigation. One is to extend the theoretical framework to accommodate fully non-convex objectives and larger-scale client deployments. Another is to integrate AMSFL with robust aggregation or privacy-preserving techniques to enhance its applicability in adversarial or privacy-critical environments. We also envision extending AMSFL to cross-device federated learning systems where computation, energy, and bandwidth constraints are jointly optimized under real-time dynamics.

\appendix

\section{Appendix: Proofs of Theoretical Results}
\label{appendix:proofs}

This appendix contains the full mathematical derivations and proofs for the theoretical results presented in Section~\ref{sec:theory_main}. For clarity, we maintain the original numbering of the theorems and propositions as stated in the main text.
\subsection{Proof of Theorem 3.1: Error Recursion of AMSFL}
\label{app:proof_recursion}

We provide a detailed derivation of the global error propagation under multi-step local SGD updates with gradient difference approximation (GDA).

\vspace{0.5em}
\paragraph{Step 1: Local model error.}

Let $w_i^{(t)}$ be the local model of client $i$ at step $t$ in round $k$, initialized as $w_i^{(0)} = w^{(k)}$, the global model. Define the local error:
\[
e_i^{(t)} := w_i^{(t)} - w^*.
\]
Under SGD update:
\[
w_i^{(t+1)} = w_i^{(t)} - \eta \nabla F_i(w_i^{(t)}),
\]
we have:
\begin{align}
e_i^{(t+1)} &= w_i^{(t+1)} - w^* = e_i^{(t)} - \eta \nabla F_i(w_i^{(t)}), \\
&= e_i^{(0)} - \eta \sum_{s=0}^{t} \nabla F_i(w_i^{(s)}). \tag{A.1.1}
\end{align}

After $t_i$ steps, the error becomes:
\begin{equation}
e_i^{(t_i)} = e_i^{(0)} - \eta \sum_{t=0}^{t_i - 1} \nabla F_i(w_i^{(t)}). \tag{A.1.2}
\end{equation}

Since $e_i^{(0)} = w^{(k)} - w^* = e^{(k)}$, and the final local model is $w_i^{(t_i)} = w^{(k)} - \eta \sum_{t=0}^{t_i - 1} \nabla F_i(w_i^{(t)})$, we define:

\begin{equation}
e_i^{(t_i)} = e^{(k)} - \eta \sum_{t=0}^{t_i - 1} \nabla F_i(w_i^{(t)}). \tag{A.1.3}
\end{equation}

\vspace{0.5em}
\paragraph{Step 2: GDA approximation of gradient drift.}

To simplify, apply first-order GDA expansion to each local gradient term:
\begin{align}
\nabla F_i(w_i^{(t)}) &= \nabla F_i(w^{(k)}) + \left[\nabla F_i(w_i^{(t)}) - \nabla F_i(w^{(k)}) \right] \\
&= \nabla F_i(w^{(k)}) + \Delta g_i^{(t)}, \tag{A.1.4}
\end{align}
where $\Delta g_i^{(t)}$ denotes the gradient deviation at local step $t$.

Substituting into Eq.~(A.1.3), we get:
\begin{align}
e_i^{(t_i)} &= e^{(k)} - \eta \sum_{t=0}^{t_i - 1} \left( \nabla F_i(w^{(k)}) + \Delta g_i^{(t)} \right) \\
&= e^{(k)} - \eta t_i \nabla F_i(w^{(k)}) - \eta \sum_{t=0}^{t_i - 1} \Delta g_i^{(t)}. \tag{A.1.5}
\end{align}

Let us define the cumulative drift term:
\begin{equation}
\Delta_i^{(t_i)} := \sum_{t=0}^{t_i - 1} \Delta g_i^{(t)}, \tag{A.1.6}
\end{equation}
then:
\begin{equation}
e_i^{(t_i)} = e^{(k)} - \eta t_i \nabla F_i(w^{(k)}) - \eta \Delta_i^{(t_i)}. \tag{A.1.7}
\end{equation}

\vspace{0.5em}
\paragraph{Step 3: Aggregation and global error propagation.}

The global model is updated by weighted averaging:
\[
w^{(k+1)} = \sum_i \omega_i w_i^{(t_i)},
\]
so the new global error is:
\begin{align}
e^{(k+1)} &= w^{(k+1)} - w^* = \sum_i \omega_i e_i^{(t_i)} \\
&= e^{(k)} - \eta \sum_i \omega_i t_i \nabla F_i(w^{(k)}) - \eta \sum_i \omega_i \Delta_i^{(t_i)}. \tag{A.1.8}
\end{align}

Taking squared norm:
\begin{align}
\|e^{(k+1)}\|^2 &= \|e^{(k)} - \eta \sum_i \omega_i t_i \nabla F_i(w^{(k)}) - \eta \sum_i \omega_i \Delta_i^{(t_i)} \|^2 \\
&= \|e^{(k)}\|^2 - 2\eta \left\langle e^{(k)}, \sum_i \omega_i t_i \nabla F_i(w^{(k)}) \right\rangle \\
&\quad + \eta^2 \left\| \sum_i \omega_i t_i \nabla F_i(w^{(k)}) + \sum_i \omega_i \Delta_i^{(t_i)} \right\|^2. \tag{A.1.9}
\end{align}

This completes the derivation of the error propagation recurrence, as stated in Theorem~\ref{thm:recursion}.
\subsection{Proof of Theorem 3.2 and 3.3: Convergence Bound of AMSFL}
\label{app:proof_residual}

We begin from the expanded error propagation derived in Appendix~\ref{app:proof_recursion}:
\[
e^{(k+1)} = e^{(k)} - \eta \sum_i \omega_i t_i \nabla F_i(w^{(k)}) - \eta \sum_i \omega_i \Delta_i^{(t_i)}.
\]
Taking squared norm:
\[
\|e^{(k+1)}\|^2 = \left\| e^{(k)} - \eta \sum_i \omega_i t_i \nabla F_i(w^{(k)}) - \eta \sum_i \omega_i \Delta_i^{(t_i)} \right\|^2.
\]

We define:
- \( A := \eta \sum_i \omega_i t_i \nabla F_i(w^{(k)}) \),
- \( B := \eta \sum_i \omega_i \Delta_i^{(t_i)} \),

and rewrite:
\[
\|e^{(k+1)}\|^2 = \|e^{(k)} - A - B\|^2.
\]

Applying the squared norm identity:
\[
\|a - b - c\|^2 = \|a\|^2 - 2\langle a, b \rangle - 2\langle a, c \rangle + \|b\|^2 + \|c\|^2 + 2\langle b, c \rangle,
\]
we obtain:
\begin{align}
\|e^{(k+1)}\|^2 = & \|e^{(k)}\|^2 - 2\eta \left\langle e^{(k)}, \sum_i \omega_i t_i \nabla F_i(w^{(k)}) \right\rangle \\
& - 2\eta \left\langle e^{(k)}, \sum_i \omega_i \Delta_i^{(t_i)} \right\rangle + \eta^2 \left\| \sum_i \omega_i t_i \nabla F_i(w^{(k)}) \right\|^2 \\
& + \eta^2 \left\| \sum_i \omega_i \Delta_i^{(t_i)} \right\|^2 + 2\eta^2 \left\langle \sum_i \omega_i t_i \nabla F_i(w^{(k)}), \sum_i \omega_i \Delta_i^{(t_i)} \right\rangle. \tag{A.2.1}
\end{align}

Now we bound the two inner products using Cauchy-Schwarz and Young's inequality.

Let us define:
- \( G_k := \|\sum_i \omega_i t_i \nabla F_i(w^{(k)})\| \),
- \( D_k := \|\sum_i \omega_i \Delta_i^{(t_i)}\| \),
- \( B := \left\langle e^{(k)}, \sum_i \omega_i \Delta_i^{(t_i)} \right\rangle \),
- \( C := \left\langle \sum_i \omega_i t_i \nabla F_i(w^{(k)}), \sum_i \omega_i \Delta_i^{(t_i)} \right\rangle \).

Applying Young's inequality with parameter \(\rho > 0\):
\[
2\eta |B| \leq 2\eta \|e^{(k)}\| \cdot D_k \leq \rho \|e^{(k)}\|^2 + \frac{\eta^2}{\rho} D_k^2,
\]
\[
2\eta^2 |C| \leq \eta^2 (G_k^2 + D_k^2).
\]

Substituting back into Eq. (A.2.1), we get:
\begin{align}
\|e^{(k+1)}\|^2 \leq & \|e^{(k)}\|^2 - 2\eta \left\langle e^{(k)}, \sum_i \omega_i t_i \nabla F_i(w^{(k)}) \right\rangle + \rho \|e^{(k)}\|^2 + \frac{\eta^2}{\rho} D_k^2 \\
& + \eta^2 G_k^2 + \eta^2 D_k^2 + \eta^2 (G_k^2 + D_k^2). \tag{A.2.2}
\end{align}

Now we group terms:
\[
\|e^{(k+1)}\|^2 \leq (1 + \rho) \|e^{(k)}\|^2 - 2\eta \left\langle e^{(k)}, \sum_i \omega_i t_i \nabla F_i(w^{(k)}) \right\rangle + \eta^2 G_k^2 (1 + 1) + \eta^2 D_k^2 \left(1 + \frac{1}{\rho} + 1 \right).
\]

Let us define:
- \( \theta := 2\eta \mu E \), where \( E := \sum_i \omega_i t_i \),
- then using strong convexity:
\[
\left\langle e^{(k)}, \sum_i \omega_i t_i \nabla F_i(w^{(k)}) \right\rangle \geq \mu E \|e^{(k)}\|^2,
\]

so:
\[
\|e^{(k+1)}\|^2 \leq (1 - \theta + \rho) \|e^{(k)}\|^2 + 2\eta^2 G_k^2 + \left(2 + \frac{1}{\rho}\right) \eta^2 D_k^2.
\]

Letting \(\theta' = \theta - \rho\) and collecting constants:
\[
\|e^{(k+1)}\|^2 \leq (1 - \theta') \|e^{(k)}\|^2 + \Delta_k, \tag{A.2.3}
\]
where:
\[
\Delta_k := 2\eta^2 G_k^2 + \left(2 + \frac{1}{\rho}\right) \eta^2 D_k^2.
\]

This completes the derivation of linearized error recursion (Eq. 3.32). Applying convergence of non-negative recurrence:
\[
\limsup_{k \to \infty} \|e^{(k)}\|^2 \leq \frac{\Delta_k}{\theta'}.
\]

This proves Theorem~\ref{thm:residual_bound}.

\subsection{Proof of Proposition 3.3: Approximation Error of GDA}
\label{app:proof_gda}

We aim to quantify the approximation error introduced by replacing the Hessian-vector product \( \nabla^2 F(w) \cdot \delta \) with a first-order gradient difference:
\[
\Delta_{\text{GDA}} := \nabla F(w + \delta) - \nabla F(w) - \nabla^2 F(w) \cdot \delta.
\]

We now prove that:
\[
\|\Delta_{\text{GDA}}\| \leq \frac{L}{2} \|\delta\|^2,
\]
where \( F : \mathbb{R}^d \to \mathbb{R} \) is twice continuously differentiable and its gradient \( \nabla F \) is \( L \)-Lipschitz, i.e.,
\[
\|\nabla F(x) - \nabla F(y)\| \leq L \|x - y\|, \quad \forall x, y.
\]

\vspace{0.5em}
\paragraph{Step 1: Taylor expansion with integral remainder.}

By the second-order Taylor theorem with integral form of the remainder, we have:
\[
\nabla F(w + \delta) = \nabla F(w) + \nabla^2 F(w) \cdot \delta + R(\delta),
\]
where:
\[
R(\delta) = \int_0^1 \left[ \nabla^2 F(w + t \delta) - \nabla^2 F(w) \right] \cdot \delta \, dt.
\]

Thus,
\[
\Delta_{\text{GDA}} = R(\delta) = \int_0^1 \left[ \nabla^2 F(w + t \delta) - \nabla^2 F(w) \right] \cdot \delta \, dt.
\]

\vspace{0.5em}
\paragraph{Step 2: Norm bound using Lipschitz continuity.}

Since \( \nabla F \) is \( L \)-Lipschitz, it follows that \( \nabla^2 F \) is also bounded in operator norm:
\[
\left\| \nabla^2 F(w + t \delta) - \nabla^2 F(w) \right\| \leq L t \|\delta\|.
\]

Therefore, for any \( t \in [0, 1] \),
\[
\left\| \left[ \nabla^2 F(w + t \delta) - \nabla^2 F(w) \right] \cdot \delta \right\| \leq L t \|\delta\|^2.
\]

Taking the integral over \( t \in [0, 1] \):
\[
\|\Delta_{\text{GDA}}\| \leq \int_0^1 L t \|\delta\|^2 dt = L \|\delta\|^2 \int_0^1 t \, dt = \frac{L}{2} \|\delta\|^2.
\]

\vspace{0.5em}
\paragraph{Conclusion.}
We conclude:
\[
\left\| \nabla F(w + \delta) - \nabla F(w) - \nabla^2 F(w) \cdot \delta \right\| \leq \frac{L}{2} \|\delta\|^2.
\]

This confirms that the gradient difference approximation introduces a second-order error, which is negligible when \( \|\delta\| \to 0 \), justifying the use of GDA in resource-limited federated settings.

\subsection{Proof of Theorem 3.4: Optimal Step Allocation under Time Constraint}
\label{app:proof_step_allocation}

We aim to solve the following optimization problem:
\begin{equation}
\begin{aligned}
\min_{\{t_i\}} & \quad \alpha \sum_i \omega_i t_i + \beta \sum_i \omega_i \cdot \frac{t_i(t_i - 1)}{2} \\
\text{s.t.} & \quad \sum_i (c_i t_i + b_i) \leq S, \quad t_i \in \mathbb{N}^+.
\end{aligned}
\label{eq:opt_step_main}
\end{equation}

Here:
- \( \alpha = 2 \eta \sqrt{\mu} G_k \): convergence-driven term;
- \( \beta = \frac{1}{2} \eta^2 L^2 G^2 \): curvature-drift penalty;
- \( c_i \): local step time on client \( i \), \( b_i \): communication delay;
- \( S \): total time budget;
- \( \omega_i \): client weight in global aggregation.

---

\paragraph{Step 1: Relax the problem to continuous domain.}

Let \( t_i \in \mathbb{R}^+ \). Define:
\[
f_i(t_i) := \alpha \omega_i t_i + \beta \omega_i \cdot \frac{t_i(t_i - 1)}{2}.
\]

Then the objective becomes convex in \( t_i \), and the constraint is linear.

We form the Lagrangian:
\[
\mathcal{L}(t, \lambda) = \sum_i f_i(t_i) + \lambda \left( \sum_i (c_i t_i + b_i) - S \right).
\]

---

\paragraph{Step 2: Take KKT conditions.}

For optimality, we require the partial derivatives:
\begin{align}
\frac{\partial \mathcal{L}}{\partial t_i} &= \alpha \omega_i + \beta \omega_i \cdot \frac{2t_i - 1}{2} + \lambda c_i = 0, \\
\Rightarrow t_i &= \frac{1}{\beta \omega_i} \left( - \alpha \omega_i - \lambda c_i + \frac{1}{2} \beta \omega_i \right).
\end{align}

Rewriting:
\[
t_i = \frac{1}{2} + \frac{1}{\beta \omega_i} \left( - \alpha \omega_i - \lambda c_i \right).
\]

To ensure \( t_i > 0 \), we require:
\[
\lambda < \frac{1}{2} \beta \omega_i - \alpha \omega_i \cdot \frac{1}{c_i}.
\]

---

\paragraph{Step 3: Structural insight and monotonicity.}

We now derive the qualitative form of the optimal allocation.

Assume all \( b_i \) are constant or pre-accounted, and consider large \( t_i \), then the cost is dominated by:
\[
f_i(t_i) \sim \frac{\beta \omega_i}{2} t_i^2, \quad \text{subject to } \sum_i c_i t_i \leq S.
\]

This leads to a standard quadratic resource allocation problem:
\[
\min \sum_i a_i t_i^2, \quad \text{s.t. } \sum_i c_i t_i \leq S,
\]
where \( a_i = \frac{\beta \omega_i}{2} \). The classical solution is:
\[
t_i^* \propto \left( \frac{1}{c_i a_i} \right)^{1/2} = \left( \frac{1}{c_i \omega_i} \right)^{1/2}.
\]

---

\paragraph{Conclusion.}

Hence, the optimal step allocation under time constraint satisfies:
\[
t_i^* \propto \left( \frac{1}{c_i \omega_i} \right)^{1/2},
\]
or, in simpler cases with uniform \( \omega_i \), \( t_i^* \propto \left( \frac{1}{c_i} \right)^{1/2} \).

This completes the proof.

\bibliographystyle{plain}

\bibliography{references}

\end{document}